\documentclass[a4paper,conference]{IEEEtran}
\pdfoutput=1
\usepackage{textcomp}
\usepackage{amsmath}
\usepackage{comment}
\usepackage{eucal}
\usepackage{textcomp}
\usepackage{amsfonts}
\usepackage{booktabs}
\usepackage{multirow}
\usepackage{textcomp}
\usepackage{graphicx}
\usepackage{breqn}
\usepackage{siunitx}
\usepackage{braket}
\usepackage{algorithm}
\usepackage{dsfont}
\usepackage{stmaryrd}
\usepackage{algpseudocode}
\newtheorem{definition}{Definition}

\author{
\IEEEauthorblockN{Simone Martone, Francesco Manigrasso, Fabrizio Lamberti and Lia Morra}
\IEEEauthorblockA{Department of Control and Computer Engineering 
Politecnico di Torino}
simone.martone@studenti.polito.it, \{francesco.manigrasso, fabrizio.lamberti, lia.morra\}@polito.it

}

\begin{document}
\title{PROTOtypical Logic Tensor Networks (PROTO-LTN) for Zero Shot Learning}
\maketitle
\begin{abstract}
Semantic image interpretation can vastly benefit from approaches that combine sub-symbolic distributed representation learning with the capability to reason at a higher level of abstraction. Logic Tensor Networks (LTNs) are a class of neuro-symbolic systems based on a differentiable, first-order logic grounded into a deep neural network. LTNs replace the classical concept of training set with a knowledge base of fuzzy logical axioms. By defining a set of differentiable operators to approximate the role of connectives, predicates, functions and quantifiers, a loss function is automatically specified so that LTNs can learn to satisfy the knowledge base. We focus here on the subsumption or \texttt{isOfClass} predicate, which is fundamental to encode most semantic image interpretation tasks. Unlike conventional LTNs, which rely on a separate predicate for each class (e.g., dog, cat), each with its own set of learnable weights, we propose a common \texttt{isOfClass} predicate, whose level of truth is a function of the distance between an object embedding and the corresponding class prototype. The PROTOtypical Logic Tensor Networks (PROTO-LTN) extend the current formulation by grounding abstract concepts as parametrized class prototypes in a high-dimensional embedding space, while reducing the number of parameters required to ground the knowledge base.
We show how this architecture can be effectively trained in the few and zero-shot learning scenarios. Experiments on Generalized Zero Shot Learning benchmarks validate the proposed implementation as a competitive alternative to traditional embedding-based approaches. The proposed formulation opens up new opportunities in zero shot learning settings, as the LTN formalism allows to integrate background knowledge in the form of logical axioms to compensate for the lack of labelled examples. PROTO-LTN was implemented in Tensorflow and is available at https://github.com/FrancescoManigrass/PROTO-LTN.git

\end{abstract}

\IEEEpeerreviewmaketitle

\makeatletter
\let\oldabs\abs
\def\abs{\@ifstar{\oldabs}{\oldabs*}}
\let\oldnorm\norm
\def\norm{\@ifstar{\oldnorm}{\oldnorm*}}
\makeatother

\newcommand{\XX}{\mathcal{X}}
\newcommand{\CC}{\mathcal{C}}
\newcommand{\FF}{\mathcal{F}}
\newcommand{\PP}{\mathcal{P}}
\newcommand{\GG}{\mathcal{G}}
\newcommand{\DD}{\mathcal{D}}
\newcommand{\LL}{\mathcal{L}}
\newcommand{\KK}{\mathcal{K}}
\newcommand{\prob}{\mathbb{P}}
\newcommand{\reals}{\mathbb{R}}
\newcommand{\Din}{D_{\mathrm{in}}}
\newcommand{\Dout}{D_{\mathrm{out}}}
\newcommand{\naturals}{\mathbb{N}}
\newcommand{\sigmoid}{\mathrm{sigmoid}}
\newcommand{\fuz}{\mathrm{FuzzyOp}}
\newcommand{\satagg}{\mathrm{SatAgg}}
\newcommand{\argmaxx}{\mathrm{argmax}}
\newcommand{\argminn}{\mathrm{argmin}}
\newcommand{\qt}[1]{``#1''}
\newcommand{\bigtimes}{\vartimes}

\section{Introduction}
\label{sec:introduction}

Despite their impressive performance when trained on large-scale, supervised datasets, deep neural networks have still difficulties generalizing to unseen categories. On the contrary, humans can leverage logical reasoning to make guesses about new circumstances, and are able to infer knowledge from few to zero examples. Recent efforts towards Neural-Symbolic (NeSy) integration \cite{de2021statistical,besold2017neural} allow to assimilate symbolic representation and reasoning into deep architectures: this entails that background knowledge, in the form of logical axioms, can be exploited during training, opening up new scenarios for settings in which labelled examples are scarce or noisy \cite{donadello2019compensating,manigrasso2021faster}. Specifically, we focus here on Logic Tensor Networks (LTNs) \cite{LTN}, a NeSy architecture that replaces the classical concept of a training set with a Knowledge Base $\mathcal{K}$ of logical axioms, ultimately interpreted in a fuzzy way, and formulates the learning objective as maximizing the satisfiability of $\mathcal{K}$. While this framework has been applied to multi-label classification problems \cite{LTN,LTN2} and object detection \cite{manigrasso2021faster}, its application to  few- and zero-shot image classification  has not yet been investigated. 

In this work, we explore this task from a NeSy perspective, and propose to integrate ideas and concepts from the few-shot learning (FSL) and zero-shot learning (ZSL) domains, namely the Prototypical Networks (PNs) \cite{prototypical} framework, within the LTN formulation. PNs define  class prototypes in a high-dimensional embedding space, so that incoming examples are assigned to the class of their nearest prototype according to some distance measure. In the LTN framework, this is achieved by representing the \texttt{isOfClass} relationship as a function of the distance between a class prototype and an object instance, thus obtaining the Prototypical Logic Tensor Network (PROTO-LTN) architecture. As the embedding space is the focus of the learning procedure, such prototypes may be also defined for classes that are not seen at training time.

The present study thus formulates a theoretical framework that achieves competitive results with respect to standard embedding-based ZSL architectures such as DEM \cite{dem}, yet offering higher degrees of flexibility. Although our analysis shows that their basic settings the two formulations are equivalent, PROTO-LTNs have greater potential in both standard and transductive ZSL. They are able to integrate in the training process prior knowledge and logical constraints from an external knowledge base, including information related to unseen classes \cite{wan2019transductive}. Hence, a NeSy formulation allows to constraint the embedding space via symbolic priors. 

The proposed framework has also potential advantages over traditional LTNs,  even outside of the FSL and ZSL settings, since classes are represented as parametrized prototypes rather than a discrete label space \cite{LTN,manigrasso2021faster}. First, representing higher-level concepts as distributed vectorized representations allows to naturally exploit the notion of distance for highlighting relationships between symbols, with semantically related symbols having similar representations \cite{goyal2020inductive}. Second, prototypes allow to ground abstract concepts in a vectorized form that can be more easily manipulated: as an example, it would be easier to define a suitable grounding for predicates that directly operate on the abstract classes, as well as their instances. Third, prototypes are more interpretable than simple labels, as their incorporation into the embedding space can be easily visualized by employing dimensionality reduction methods, such as t-SNE \cite{van2008visualizing}.

The rest of the paper is organized as follows. In Section \ref{sec:related}, we place the present work in the context of the related literature, and provide a background on LTNs. In Section \ref{sec:protoltn}, we describe a simple theoretical scheme to assimilate PNs into a LTN for classification purposes (PROTO-LTN),  both in the FSL and ZSL scenarios. Then, in Sections \ref{sec:experimental} and \ref{sec:results}, we examine the behavior of the model in the Generalized Zero-Shot-Learning (GZSL) task on common benchmark datasets. Finally, in Section \ref{sec:conclusion}, we discuss 
conclusions and future works.

\section{Related work}
\label{sec:related}
\subsection{Neural-symbolic AI in Semantic Image Interpretation}
Research on how to combine connectionist and symbolic approaches has flourished in the past few years \cite{LTN,yu2021survey}, with several applications in semantic image interpretation and visual query answering \cite{LTN,manigrasso2021faster,donadello_semantic_image_interpretation,donadello2019compensating,yi2018neural,vedantam2019probabilistic,li2021calibrating}. Among the plethora of compositional patterns that have been proposed \cite{van2021modular,yu2021survey}, the present work follows two main principles: knowledge representation (in the form of first order logic) is embedded into a neural network, which in turn allows to constrain the search space by leveraging explicit (and human-interpretable) domain knowledge as a symbolic prior. This latter property is extremely useful in ZSL, in which some external source of information is exploited to offer an abstract description of the classes in lieu of providing training examples. On the other hand, compared to approaches based on Inductive Logic Programming (such as \cite{yi2018neural}), in which perception and reasoning are performed by separate modules, LTNs provide tighter integration between the two subsystems.

\subsection{Logic Tensor Networks}

LTNs have proven effective in higher-level image interpretation tasks, such as object detection and scene graph construction \cite{donadello_semantic_image_interpretation,LTN}. Donadello et al. applied them for scene relationship detection in a zero shot setting, showing how prior knowledge can compensate for the lack of supervision \cite{donadello2019compensating}. 

In the LTN framework, the term \textit{grounding} denotes the interpretation of a First Order Language  into a subset of the $\mathbb{R}^{n}$  domain \cite{LTN}. It defines a collection of \textit{terms} (objects) and \textit{formulas} described in a \textit{Knowledge base} $\mathcal{K}$. For instance, to express the friendship between two terms defined as \textit{Alice} and \textit{Bob}, we can use the predicate \texttt{friend\_of}:
\begin{align*}
   \phi_{1} = \texttt{friend\_of} (Alice, Bob)  \wedge  \texttt{friend\_of}(Bob, Alice) 
\end{align*}
At the same time, we can specify formulas defining general properties, such as the symmetric nature of the friendship relationship within a specific \textit{domain}:
\begin{align*}
    \phi_2 = \forall \, x, y \, (\texttt{friend\_of}(x,y) \Rightarrow \texttt{friend\_of}(y,x))
\end{align*}

Adopting Real Logic, both formulas and terms are \textit{grounded} (interpreted) into a scalar value in the [0,1] interval. Specifying the grounding function $\mathcal{G}$, which maps terms and formulas into such real-valued features, generates a complete definition of a theory. Given a set of terms, aggregate formulas can be defined by approximating unary, binary or quantifiers connectives in fuzzy logic using suitable differential operators.

In semantic image interpretation tasks, terms (objects) are typically grounded by features computed by a pre-trained convolutional neural network; it is also possible to jointly train the convolutional backbone and the LTNs in an end-to-end fashion \cite{manigrasso2021faster}. Predicates symbols $p \in \mathcal{P} $ are grounded by a function $\mathcal{G}\left(D(p)\right) \rightarrow[0,1]$. A typical predicate in semantic image interpretation is the $\texttt{isOfClass}$ one, which represents the probability that a given object belongs to class $c$. 

In conventional LTNs \cite{LTN,donadello_semantic_image_interpretation,manigrasso2021faster}, predicates are typically defined as the generalization of the neural tensor network: 
\begin{align}
\label{eqn:predicate}
\mathcal{G}\left( \mathcal{P} \right)( \mathbf{v}) = \sigma\left(\mathit{u_{P}^{T}}\tanh\left( \mathbf{v_{T}} W_{P}^{[1:k]} \mathbf{v} + V_{P} \mathbf{v} + \mathit{b_{p}} \right) \right)
\end{align}
where $\sigma$ is the sigmoid function, $W[1:k] \in \mathbb{R}^{k \times mn \times mn}$, $V_{p} \in \mathbb{R}^{k \times mn}$, $u_{p} \in \mathbb{R}^{k}$, and $ b_{p} \in \mathbb{R} $  are learnable tensors of parameters. For multi-class problems, the sigmoid function could be substituted by a softmax layer to enforce mutual exclusivity \cite{LTN}.

This grounding requires to add an additional predicate for each class (e.g., \texttt{isDog}, \texttt{isPerson}, etc.), which is embedded into a tensor network with separate weights. Additionally, since class symbols are not grounded, predicates can only be defined for object instances, which rapidly leads to very large knowledge bases when background logical axioms need to be imposed. On the contrary, our proposed grounding does not require additional model parameters, or in any case limits them to a small set which is shared among all $\texttt{isOfClass}$ predicates. Furthermore, it encodes abstract classes as parametric objects that live in the same embedding space as their instances, and can be used to establish relationships with other objects (e.g., macro-category relationships).  This formulation thus supports more efficient and compact representations.

The \textit{best satisfability} problem, which is the optimization problem underlying LTNs, consists in determining the values of $\Theta^{*}$ that maximize the truth values of the conjunction of all formulas $\phi \in \mathcal{K}$:
\begin{align}
\label{eq:best_sat_2}
\Theta^{*}= argmax_{\Theta} \hat{\mathcal{G}}_{\theta}\left(\bigwedge_{\phi \in \mathcal{K}} \phi \right) - \lambda|| \Theta ||_{2}^{2}
\end{align}
where $ \lambda|| \Theta ||_{2}^{2}$ is a convenient regularization term.

\subsection{Zero-shot learning}

In zero-shot learning, a learner must be able to recognize objects from test classes, not seen during training, by leveraging some sort of description, most commonly a vector of semantic attributes \cite{awa2}. In this paper, we target the Generalized zero-shot learning (GZSL) scenario, in which both seen and unseen classes appear at test time \cite{awa2}. State-of-the-art techniques for ZSL classification typically fall within two categories \cite{awa2,dem}: \textit{embedding-based} and \textit{generative-based}.

Embedding-based models \cite{dem,relation,vse,devise} compare semantic characteristics (e.g., attributes) and visual characteristics (usually taken from a pre-trained convolutional neural network) by (learning a) mapping to a common embedding space. Mapping the semantic space to the more compact visual feature space, rather than the opposite, alleviates the so-called hubness problem and facilitates separation between classes \cite{dem}. Standard embedding-based models are completely agnostic to any information about the test set: neither examples (even unlabelled), nor class attributes are assumed to be available at training time.  Although based on a NeSy formulation, the proposed PROTO-LTN approach can be regarded as an embedding-based technique, as semantic concepts and visual features are mapped onto a common embedding space. 

Embedding-based models tend to be naturally biased towards seen classes. To alleviate this problem, generative models were proposed with the purpose of learning a conditioned probability distribution for each class, and thus generate artificial examples of unseen classes  \cite{Verma_2018_CVPR,gdan,robustbidirectional}.  A conventional classifier is trained by utilizing both the true and the generated examples. Although impressive results, especially in a GZSL context, can be achieved by taking advantage of this machinery, reduced flexibility with respect to embedding methods is entailed, as unseen classes need to be defined, so that a number of corresponding examples can be artificially synthesized. PROTO-LTNs are thus best compared with other embedding-based models, although nothing prevents them from being trained on, or combined with, generative methods.

\section{PROTOtypical Logic Tensor Networks}
\label{sec:protoltn}

First, we introduce the basic notations related to prototypical networks in the FSL (Section \ref{ssec:pn_fsl}) and ZSL (Section \ref{ssec:pn_gzsl}) settings \cite{prototypical}. Then, in Sections \ref{ssec:protoltn-fsl} and \ref{ssec:protoltn-gzsl}, we build on these concepts and show how the PROTO-LTN training cycle is constructed by substituting the original model with a grounded $\mathcal{K}$, and the original loss with a best satisfiability problem. 

\subsection{Prototypical Networks: the FSL setting}
\label{ssec:pn_fsl}
A $N$-way-$K$-shot FSL scenario is supposed, in which a classifier is asked to discriminate the right class among $N$ choices, while having the chance to observe $K$ examples per class \cite{few1,few2,few3}. More specifically, the labelled examples are referred to as the \textit{support} examples, whereas the unlabeled ones as the \textit{query} examples.

The underlying assumption that it exists an embedding space in which elements of different classes are well-scattered, and that it can be mathematically translated into an embedding function $f_\theta$ whose parameter $\theta$ must be inferred, acting as a mapping 
\begin{align}
    f_\theta : \reals^D \to \reals^M.
    \label{eq:ftheta}
\end{align}

In Eq. \ref{eq:ftheta}, $D$ and $M$ are, respectively, the dimensions of the input space and  of the embedding space. Thus, for an example $x$, $f_\theta(x)$ is the corresponding embedding. 

In FSL, a \textit{prototype} for class $n$ is obtained as the mean embedding of the $K$ support examples of class $n$ at train time:
    \begin{align}
        \label{batch_prototype}
        p_n = \frac{1}{K} \sum_{\substack{(x^{\Tilde{S}},y^{\Tilde{S}}) \in \Tilde{S} \\ \text{s.t. } y^{\Tilde{S}} = n}} f_\theta (x^{\Tilde{S}}).
    \end{align}
Class prototypes thus need to live in the embedding space, as they embody average features shared by elements of the class they represent. %
At \textit{training time},  $\theta$ is optimized so that the distance between each prototype and the elements of its class is minimized, while the distance between different prototypes is maximized. Finally, classification at \textit{testing time} is performed by assigning each query sample to its nearest prototype.

At testing time, a support set is at disposal of $N_S$ labeled examples $S = \{ (x^S_1,y^S_1), ..., (x^S_{N_S},y^S_{N_S}) \}$, where each $x^S_i \in \reals^D$ is the feature vector of an example, and $y^S_i \in C \subset \mathbb{N}$ is the corresponding label. Assuming a $N$-way-$K$-shot scenario, exactly $K$ support examples are available for each of the $N$ classes. A query set $Q = \{ x^Q_1, ..., x^Q_{N_Q} \} $ of $N_Q$ unlabeled examples is thus supplied, and the task is to correctly assort the examples into their classes. The elements from the query set $Q$ belong to the same domain as those from the support set $S$. 

At training time, it could be impossible to know which classes will the testing scenario yield. In other words, a support set $S$ is not accessible in advance. To cope with that, a training set $T = \{ (x^T_1, y^T_1), ..., (x^T_{N_T}, y^T_{N_T}) \}$ is chosen that reflects the best prior information  possessed about the testing scenario, with labels $y_i^T \in C_T \subset \mathbb{N}$ and $|C_T| = N_T$ classes which can coincide or outnumber them ($N_T \ge N$). In other words, it is possible that $C \cap C_T \neq \emptyset$, but it cannot be said in advance. Then, \textit{fake} support and query sets $\Tilde{S} \subset T$ and $\Tilde{Q} \subset T$ are extracted to mimic the testing scenario and instruct the model to learn accordingly.

\subsection{Prototypical networks: the ZSL setting}
\label{ssec:pn_gzsl}
In ZSL, one does not dispose of labelled examples for all classes. Instead, it is  assumed that $N$ abstract vectors denoted as $\{ a^{(1)}, a^{(2)}, ..., a^{(N)} \}$, with $a^{(n)} \in \reals^A$,  encode the characteristics of all $N$ classes.

As in FSL, at training time one takes advantage of a set $T = \{ (x^T_1, y^T_1), ..., (x^T_{N_T}, y^T_{N_T}) \}$ of labelled examples from classes $y^T_i \in C_T \subset \mathbb{N}$, where it is preferably  $|C_T| = N_T \ge N = |C|$. The training cycle remains unchanged in the ZSL case, but class prototypes are defined differently:  
\begin{itemize}
    \item the embedding for a query example $x^Q$ is still  obtained as $f_\theta(x^Q)$, where $f_\theta: \reals^D \to \reals^M$;
    \item the prototype for class $n \in C$ is extracted as $p_n = g_\theta(a^{(n)})$ via a separate embedding function $g_\theta: \reals^A \to \reals^M$, which maps the semantic attribute space to the common embedding space.
\end{itemize}

\subsection{PROTO-LTN: the FSL scenario}
\label{ssec:protoltn-fsl}

\begin{figure*}[tbh!]
\centering
	\includegraphics[width=0.8\textwidth]{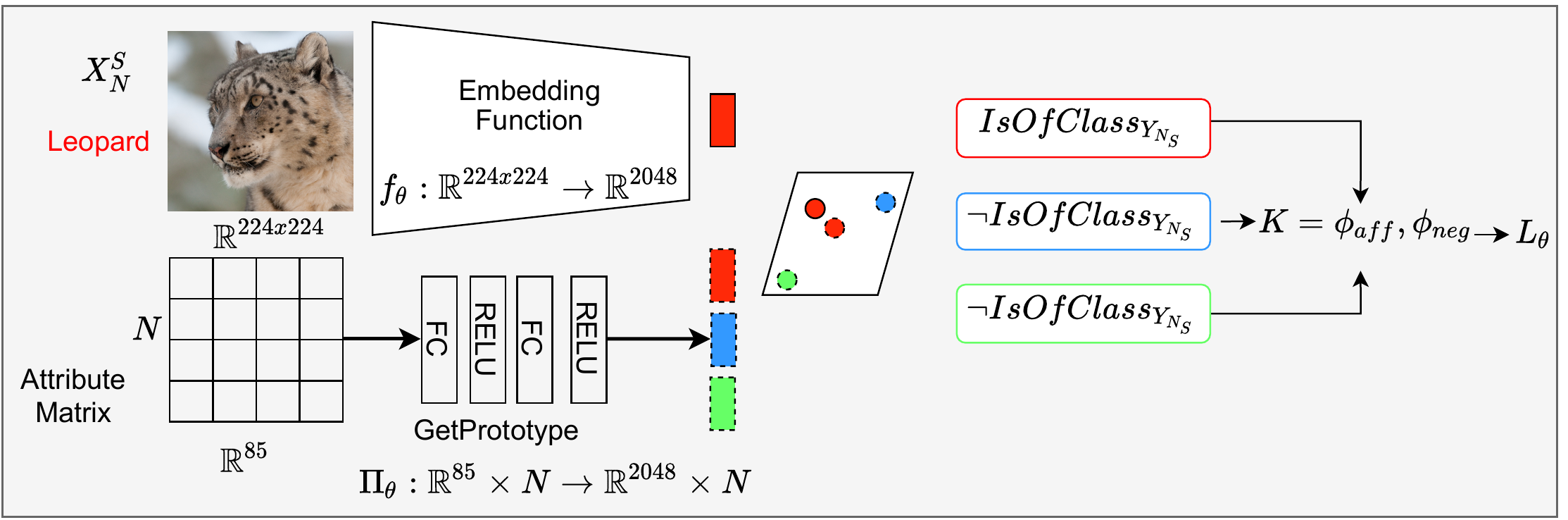}
	\caption{Proto-LTN architecture for ZSL classification. The architecture is composed of a convolutional features extractor and an attribute encoder. The two branches allow to map semantic and visual features in a common embedding space. The \texttt{isOfClass} predicate aims to minimize the distance between instances (solid line circles) and class prototypes (dashed line circles) based on affirmative and negative formulas embedded in the knowledge base $\mathcal{K}$. At train time, the loss function maximizes the satisfiability (truth value) of all formulas in $\mathcal{K}$. 
	} 
	\label{Figure:PROTO-LTN}
\end{figure*}

The overall architecture of PROTO-LTN, when tailored to the ZSL scenario, is illustrated in Fig. \ref{Figure:PROTO-LTN}. The input image embeddings are extracted from a convolutional neural network, while attribute vectors are mapped into the embedding domain through an embedding function. In this section, details about the definition of the grounding of the constant, variables, functions and predicates are given. Then, the Knowledge Base $\mathcal{K}$ which encodes our learning problem is defined. 

\subsubsection{Groundings terms}
Within a single training episode, a batch of training samples is selected in the form of fake support $\Tilde{S}$ and query $\Tilde{Q}$ sets. Groundings for variables and their domain $D$ (not learnable) can be defined as
\begin{eqnarray}
    \GG(q) &= \braket{x^{\Tilde{Q}}_{1},...,x^{\Tilde{Q}}_{N_{\Tilde{Q}}}}, \\
    \GG(q_l) &= \braket{y^{\Tilde{Q}}_{1},...,y^{\Tilde{Q}}_{N_{\Tilde{Q}}}},\\
    \GG(q_e) &= \GG(\texttt{getEmbedding}(q)) \\
    &= \braket{f_\theta(x^{\Tilde{Q}}_{1}),...,f_\theta(x^{\Tilde{Q}}_{N_{\Tilde{Q}}})}, \\
    \GG(s) &= \braket{x^{\Tilde{S}}_{1},...,x^{\Tilde{S}}_{N_S}}, \\
    \GG(s_l) &= \braket{y^{\Tilde{S}}_{1},...,y^{\Tilde{S}}_{N_S}}, \\
    \GG(p), \, \GG(p_l) &= \GG(\texttt{getPrototypes}(s,s_l)) \\
    &= \Pi_\theta(\GG(s, s_l) ) \\
    &= \Pi_\theta(\braket{(x^{\Tilde{S}}_{1},y^{\Tilde{S}}_{1}),...,(x^{\Tilde{S}}_{N_S},y^{\Tilde{S}}_{N_S})}),
\end{eqnarray}
where $q$ are the query examples ($ D(q) = \texttt{features}$), $q_l$ are the corresponding labels ($D(q_l) = \texttt{labels}$), and $q_e$ are their embeddings ($ D(q_e) = \texttt{embeddings}$). Conversely, $s$ are the examples in the support set ($D(s) = \texttt{features}$) and $s_l$ their labels. Finally, $p$ and $p_l$ are the prototypes and their labels, respectively, with $D(p) = \texttt{embeddings}$ and $D(p_l) = \texttt{labels}$. 

\subsubsection{Grounding functions and predicates}
PROTO-LTNs are based on two functions ($\texttt{getEmbedding}$  and $\texttt{getPrototypes}$) and the $\texttt{isOfClass}$ predicate. 

$\texttt{getEmbedding}$ is a conventional LTN function which maps image features into the embedding space, hence $\Din(\texttt{getEmbedding}) = \texttt{features}$ to \\ $\Dout(\texttt{getEmbedding}) = \texttt{embeddings}$.

The   \texttt{getPrototypes} function, with $\Din(\texttt{getPrototypes}) = \texttt{features} \times \texttt{labels}$ and \\ $\Dout(\texttt{getPrototypes}) = \texttt{embeddings} \times \texttt{labels}$, returns labelled prototypes given a support set of labelled examples.  Each prototype is in fact a function of all support points belonging to the same class, as defined in Eq. \ref{batch_prototype}. It is defined as a \textit{generalized} LTN function, which accepts as input multiple instantiations of variables (and hence multiple domains). A more formal definition is given in Appendix A. %

Groundings for both functions are defined as:
\begin{align}
    \GG(\texttt{getEmbedding}) &= f_\theta, \\
    \GG(\texttt{getPrototypes}) &= \Pi_\theta,
\end{align}
where $f_\theta: \reals^D \to \reals^M$ defines the embedding function, whereas
\begin{align}
    \Pi_\theta: \bigcup_{l=1}^{\infty}  \bigtimes_{m=1}^l \mathbb{R}^D \times \naturals \to \bigcup_{l=1}^{\infty}  \bigtimes_{m=1}^l   \mathbb{R}^M \times \naturals
\end{align}
accepts as input a list of $N_S$ labelled support examples, i.e., an element of $(\reals^D \times \naturals)^{N_S}$, and returns a list of labelled prototypes for all the $\Tilde{N}$ classes seen in the support set, or an element of $(\reals^M \times \naturals)^{\Tilde{N}}$. Additional details are given in Appendix A. %

The \texttt{isOfClass} predicate for class $n$ $\in C$ is grounded as:
\begin{align}
    \GG(\texttt{isOfClass}) = e^{-\alpha \, d(\cdot, \cdot)^2} \label{squared_euclidean},
\end{align}
where $\alpha$ is a hyperparameter and $d$ is a measure of distance. $\GG(\texttt{isOfClass}): \reals^M \times \reals^M \to [0,1]$; $\GG(\texttt{isOfClass})$takes the value of $ 1$ when the distance from the class prototype $d(\cdot, \cdot) $ is $ 0 $. In our formulation the Euclidean distance squared is adopted, as in DEM \cite{dem}. Alternatively, parametric similarity functions could be used: 
\begin{align}
     \GG''(\texttt{isOfClass}) &= \sigma_{\theta}( \text{Concatenate}[ \cdot, \cdot ] ). \label{parametric_similarity}
\end{align}
where $\sigma_{\theta}$ could be a MLP with output sigmoid activation. This formulation is closer to that of Relation Networks \cite{relation}.

\subsubsection{Knowledge Base}
 $\KK$ represents our knowledge about the formulated problem and is updated at each training episode based on the current fake support set. $\KK = \{ \phi_{\text{aff}}, \phi_{\text{neg}} \}$ contains two aggregations of formulas which specify that each query item is a positive example for its class, and a negative one for all the others:

\begin{dmath}
\phi_{\text{aff}} = \forall \text{Diag}(q_e,q_l) (\forall\text{Diag}(p,p_l): {q_l=p_l} 
( \texttt{isOfClass}(q_e,p) ) ), \label{affirmation_fewshot} 
\end{dmath}
\begin{dmath}
\phi_{\text{neg}} = \forall \, \text{Diag}(q_e,q_l) \, ( \forall \, \text{Diag}(p,p_l): { q_l \neq p_l }\, (\lnot \texttt{isOfClass}(q_e,p) ) ). \label{negation_fewshot}
\end{dmath}
We have exploited both Diagonal Quantification and Guarded Quantifiers, whose formal definition can be found in \cite{LTN}.

PROTO-LTN is trained by maximizing the satisfiability

\begin{equation}
        \mathcal{L}^{\text{ep}} = 1 -\left(\bigwedge_{\phi \in \mathcal{K}} \phi \right)  = -\GG(\phi_{\text{aff}}) - w_{\text{n}} \, \GG(\phi_{\text{neg}}),
\end{equation}
where the weight $w_{\text{n}}$ reflects the expectation that negations play a less discriminative role than affirmation in classification. In our experiments, we set $w_{\text{n}}=0$ and consider only $\phi_{\text{aff}}$, leaving exploration of this hyper-parameter to future work. 

By introducing an aggregation function \cite{LTN,van2008visualizing}, we obtain
\begin{dmath}
{\mathcal{L}^{\text{ep}} =  \bigg( -\log(\GG(\phi_{\text{aff}})\big)^{\frac{1}{p_{\text{agg}}}}) +  w_{\text{n}}\big(1-\GG(\phi_{\text{n}})\big)^{\frac{1}{p_{\text{agg}}}} \bigg)^{p_{\text{agg}}}}
\label{loss_sat}
\end{dmath}
where $\GG(\phi_{\text{aff}})$ is implemented through the generalized product $p$-mean operator and $\GG(\phi_{\text{neg}})$ with the generalized mean operator $A_{pM}$:
\begin{equation*}
\small
\begin{aligned}
\label{prod_mean}
    A_{pPR}(\tau_1,...,\tau_n) = \bigg(  \prod_{i=1}^n \tau_i \bigg)^{\frac{1}{p_\forall}},
\end{aligned}
\begin{aligned}
    A_{pM}(\tau_1,...,\tau_n) = \bigg( \frac{1}{n} \sum_{i=1}^n \tau_i^p \bigg)^{\frac{1}{p_\forall}}.
\end{aligned}
\end{equation*}

It should be noticed that the choice of $p_{agg}$ does not need to coincide with that of $p_\forall$ for quantification, and both hyper-parameters need to be tuned experimentally.

When optimizing a positive quantity, a common practice consists in optimizing its logarithm: the product between similarities takes a more desirable form when $A_{pPR}$ is used as the aggregation operator for $\forall$. Unfortunately, one does not obtain an equally appealing expression for $\phi_{\text{neg}}$.

 If a squared Euclidean distance is used as similarity measure and the negation weight $w_{\text{n}}$ is set to 0,  one obtains the same formulation of the loss function of DEM \cite{dem}, up to a scaling constant:
    \begin{align}
        \mathcal{L}^{\text{ep}} &= -\log \Bigg( e^{-\frac{\alpha}{p_\forall} \, \big( \sum_{n \in \Tilde{C}} \, \sum_{\substack{(x^{\Tilde{Q}},y^{\Tilde{Q}}) \in \Tilde{Q} \\ \text{s.t. } y^{\Tilde{Q}} \neq n}} \, d(f_\theta(x^{\Tilde{Q}}), p_n)^2 \big)  } \Bigg) \nonumber \\
        &= \frac{\alpha}{p_\forall} \, \Big( \sum_{n \in \Tilde{C}} \, \sum_{\substack{(x^{\Tilde{Q}},y^{\Tilde{Q}}) \in \Tilde{Q} \\ \text{s.t. } y^{\Tilde{Q}} \neq n}} \, d(f_\theta(x^{\Tilde{Q}}), p_n)^2 \Big).
        \label{log_loss_prod_no_negation}
    \end{align}

\begin{algorithm}

\small
\caption{PROTO-LTN - GZSL Training procedure}\label{alg:cap}
\begin{algorithmic}

\Function {Train  }{}
\State  Input $\leftarrow$  $q$ Training Images
\State  Input $\leftarrow$  $q_{l}$ Training label
\State  Input $\leftarrow$  $a$ Semantic attribute set
\State  Input $\leftarrow$  $a_{l}$ Semantic attribute label

\For {  $i $ $in$ $N_{TrainingSteps}$} 

\State  $q_{e_i} \leftarrow $  \texttt{getEmbedding}($q$)

\State $a_{i}$,$a_{l_i} \leftarrow $  \texttt{getAttributes}($a$)

\State $p_{i}, p_{l_i} \leftarrow \texttt{getPrototypes}(a_{i},a_{l_i})$

\State $\phi_{\text{aff}}$ =  $\forall \text{Diag}(q_{e_i},q_{l_i}) (\forall\text{Diag}(p_{i},p_{l_i}): {q_{l_i}=p_{l_i}} 
( \texttt{isOfClass}(q_{e_i},p_i) $
\State $\phi_{\text{n}}$ =  $\forall \, \text{Diag}(q_{i},q_{l_i}) \, ( \forall \, \text{Diag}(p_{i},p_{l_i}): { q_{l_i} \neq p_{l_i} }\, (\lnot \texttt{isOfClass}(q_{e_i},p_i))) $

\State $\bigg( \big(\log((\GG(\phi_{\text{aff}})\big)^{\frac{1}{p_{\text{agg}}}})))+  w_{\text{n}}\big(1-\GG(\phi_{\text{n}})\big)^{\frac{1}{p_{\text{agg}}}} \bigg)^{p_{\text{agg}}} $

\State $computeGradient(\mathcal{L}^{\text{ep}})$
\State $updateGradient$
\EndFor
\EndFunction
\end{algorithmic}
\begin{algorithmic}
\Function {Test  }{}
\State  Input $ \leftarrow $  $q$ Test Images
\State  Input  $ \leftarrow $   $a$ Semantic attribute set

\State  $q_{e}$  $ \leftarrow $  \texttt{getEmbedding}($q$)
\State $a$,$a_{l}$  $ \leftarrow $ \texttt{getAttributes}($a$)

\State $p, p_{l} \leftarrow$  \texttt{getPrototypes}($a$,$a_{l}$)
\For{ $i $ $in$ len($q_{e}$)}
\For{ $j $ $in$  len($p$)}
\State $prediction_{i} \leftarrow \texttt{isOfClass}(q_{e_i},p_j) $

\EndFor

\EndFor
\EndFunction
\end{algorithmic}

\end{algorithm}

\subsection{PROTO-LTN: the GZSL scenario}
\label{ssec:protoltn-gzsl}
The GZSL setting is analogous to the FSL setting, with the main difference lying in how prototypes are defined and calculated. No generalized LTN functions are needed for the GZSL case. 
Computations for a training epoch are reported in Algorithm  \ref{alg:cap}.

Since only one semantic vector $a^{(n)}$ is given for each class $n$, there is a 1-to-1 correspondence between elements of the support set and prototypes. The latter are embodied by the semantic embedding function $g_\theta: \reals^A \to \reals^D$ obtaining as the feature space the common embedding space. We just define \texttt{getPrototypes} as a conventional LTN function, whose grounding is        $\GG(\texttt{getPrototypes}) = g_\theta.$ Conversely, nothing changes for the query map \texttt{getEmbedding}.

\section{Experimental Settings}
\label{sec:experimental}

Experiments were conducted in both ZSL and GZSL settings on the {Awa2} (Animals with Attributes) \cite{awa2}, {CUB} \cite{CUB}, {aPY} (Attribute Pascal and Yahoo)\cite{aPY} and {SUN} (Scene Understanding) \cite{SUN}  benchmarks. For all datasets, image encodings, attributes and splits were collected from the original benchmark \cite{awa2}.

\setlength{\tabcolsep}{2pt} %
\renewcommand{\arraystretch}{1} %
\begin{table*}[htb]
\centering
\caption{
For PROTO-LTN, we show mean $\pm $ standard deviation and maximum (in parenthesis) performance.  $\text{TOP1}^{\text{ZSL}}$~(T1),  $\text{TOP1}^{\text{GZSL\_UNSEEN}}$~(U), $\text{TOP1}^{\text{GZSL\_SEEN}}$~(S) and    $\text{H}^{\text{GZSL}}$~(H) are always obtained on the proposed split (PS ) of Awa2, CUB, aPY and SUN classes, as described in \cite{awa2}. $\dagger$ assumes a transductive ZSL setting. Best performances are reported in bold.
}
\label{table:1}
  \begin{tabular}{l|SSSS|SSSS|SSSS|SSSS}

    \toprule
    \multirow{2}{*}{Method} &
      \multicolumn{4}{c}{Awa2} &
      \multicolumn{4}{c}{CUB } &
      \multicolumn{4}{c}{APY } &
      \multicolumn{4}{c}{SUN } \\
      & {T1} & {U} & {S} & {H}  & {T1} & {U} & {S} & {H} & {T1} & {U} & {S} & {H} & {T1} & {U} & {S} & {H} \\
      \midrule
    SYNC {(2016)} \cite{sync} & 46.6 & 10.0 & 90.5 & 18.0 &55.6 & 11.5 & \textbf{70.9} & 19.8 &{-}&{-}&{-}&{-}& 56.3 & 7.9 & 43.3 & 13.4\\
    Relation Net {(2017)}\cite{relation} & 64.2 & 30.0 & 93.4 & 45.3& 55.6 & 38.1& 61.1& 47 &{-}&{-}&{-}&{-}&{-}&{-}&{-}&{-} \\ 
    PrEN$^\dagger$ (2019) \cite{pren} & 74.1 & 32.4 & 88.6 & 47.4 &66.4 &35.2 &55.8 &43.1& {-}&{-}&{-}&{-}& \textbf{62.9} & \textbf{35.4} & 27.2 &\textbf{ 30.8}\\
    VSE {(2019)}  \cite{vse} & \textbf{84.4} & \textbf{45.6} & \textbf{88.7} & \textbf{60.2}& \textbf{71.9} & \textbf{39.5} & 68.9 & \textbf{50.2}  &\textbf{65.4} & \textbf{43.6} & \textbf{78.7} & \textbf{56.2} &{-}&{-}&{-}&{-} \\
      
     DEM{ (2017)} \cite{dem} & 67.1  & 30.5 & 86.4 & 45.1 & 51.7 & 19.6 & 57.9 & 29.2 & 35.0 & 11.1 & 75.1 & 19.4 & 61.9 & 20.5 & 34.3 & 25.6\\ \hline
     
     PROTO-LTN  & 67.6   & 32.0   & 83.7  & 46.2 & 48.8   &20.8  & 54.3  & 30.0    & 35.0    & 17.1   & 66.2  & 27.21 & 60.4   & 20.4   & \textbf{36.8}   & 26.2 \\
    &{\textpm 1.1 }&{\textpm 1.3  }&{\textpm 0.3 }&{\textpm 1.3 }&{\textpm 1.2  }&{\textpm 2.6  }&{  \textpm 1.1   }&{\textpm 3.0   }&{\textpm 3.1   }&{\textpm 2.0  }&{\textpm 5.1 }&{ \textpm 2.9 }&{ \textpm 2.5   }&{\textpm 1.0  }&{\textpm 4.4  }&{\textpm 1.9}\\
    &{(70.8)}&{(34.8)}&{(84.3)}&{(49.1)}&{(50.3)}&{(23.4)}&{(55.7)}&{(33.0)}&{(38.6)}&{(19.4)}&{(70.7)}&{(30.0)}&{(62.1)}&{(22.15)}&{(39.9)}&{(28.0)}\\

    \bottomrule
  \end{tabular}
\end{table*}

The entire architecture is composed of two different blocks: the image visual encoder and the semantic encoder. The embedding function $f_\theta$  is composed by a ResNet101  \cite{resnet101} embedding model, pretrained on ImageNet \cite{imagenet} and kept frozen, which converts an image $I$ into a vector $\mathbf{x} \in \reals^M$, where $M=2048$. This setting is maintained in all experiments with all datasets.

Semantic vectors are encoded in the embedding space via a function $g_\theta$, which consists of two fully connected layers (FC) with ReLU activation function, initialized by a truncated normal distribution function. We set the hyper-parameter aggregations to $p_{agg}=1$ and $p_\forall=2$, also taking into account preliminary experiments on Awa2 \cite{awa2}.

The framework was implemented in Tensorflow based on the LTN package \cite{LTN,ltnpackage}. Experiments were conducted on a workstation equipped with an Intel® Core™ i7-10700K CPU and a RTX2080 TI GPU. All networks were trained for 30 epochs with Adam optimizer and batch size 64. Hyper-parameters (learning rate, $\alpha$ and regularization term $\lambda$) were optimized separately for each dataset. Details are reported in Appendix B.  %
Standard performance metrics for GZSL were used as defined in \cite{awa2}. Mean and standard deviation were calculated by repeating each experiment three times.

\section{Results}
\label{sec:results}
\begin{figure}[htb]

	\includegraphics[scale=0.3]{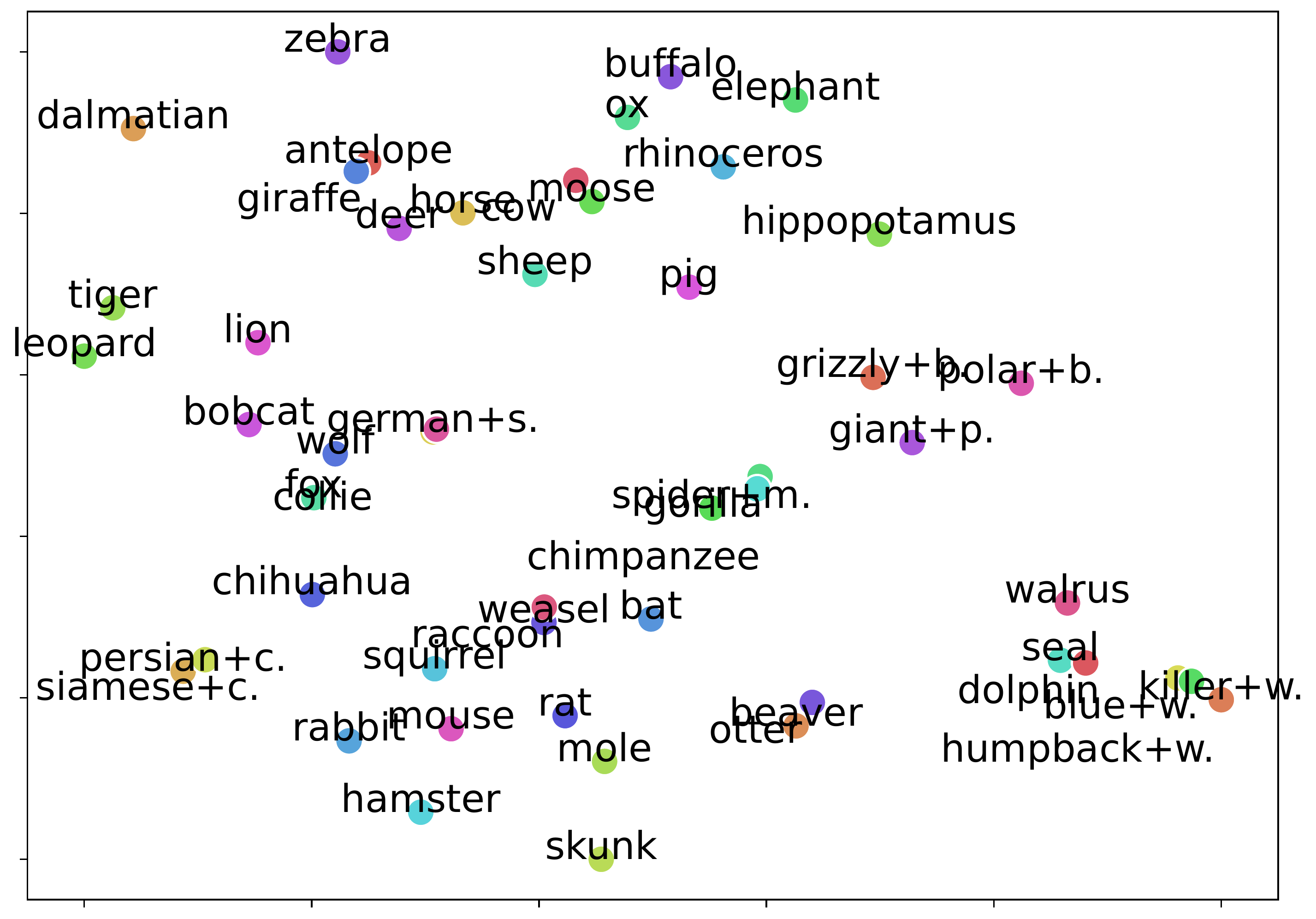}
	\caption{t-SNE visualization of class prototypes for the Awa2 dataset.  } 
	\label{Figure:TsnetEmbedding}
\end{figure}

PROTO-LTN results are reported in Table \ref{table:1}, along with those for comparable  embedding-based methods. Fig. \ref{Figure:TsnetEmbedding} illustrates the embedding space with highlighted class prototypes. 

As expected based on our analytical analysis, experimental performance is competitive with respect to most embedding-based techniques, in particular DEM \cite{dem} and Relation Net \cite{relation}, which rely on similar assumptions and the same input as the current PROTO-LTN implementation. As shown in Section \ref{ssec:protoltn-fsl}, under certain conditions the PROTO-LTN loss is equivalent to that of DEM, up to a scaling constant, albeit with different regularization terms. We outperform DEM on unseen classes for all experimental benchmarks: this entails that the proposed formulation is a strong basis for a novel, NeSy approach to the GZSL task. 

Our method is outperformed by VSE, which relies on a different strategy to compute visual feature embeddings.
A semantic loss allows to align the embedding space with part-feature concepts provided by a semantic oracle. Since the latter relies on an external knowledge base, it contains concepts beyond the available semantic vector $\{ a^{(1)}, a^{(2)}, ..., a^{(N)} \}$. This is especially advantageous in benchmarks like aPY, in which attributes are noisy and not visually informative \cite{vse}. This is a limitation of our current experiments, but not intrinsic to PROTO-LTNs. Indeed, $\mathcal{K}$ can be extended to include part-of relationships between concepts, and previous works have shown how these relationships can be leveraged to impose symbolic priors during learning, e.g., in object detection \cite{manigrasso2021faster,donadello_semantic_image_interpretation}. However, the LTN formalism needs to be further extended to align part-based concepts with their visual groundings in an unsupervised fashion.

\section{Conclusions and Future works}
\label{sec:conclusion}

We introduced PROTO-LTN, a novel Neuro-Symbolic architecture which extends the classical formulation of LTN borrowing from embeddings-based techniques. Following the strategy of PNs, we entirely  focus on learning embedding functions (such as $f_\theta$ and $g_\theta$), implying that class prototypes are obtained ex-post, based on a support set. These methods are robust to noise, an essential property in FSL, and provide a scheme to embed both examples (images) and class prototypes in the same metric space. This is a key property in the context of LTNs, because it enables different levels of abstraction: one can either state something about a particular example, or about an entire class, as prototypes can be viewed as parametrized labels for classes. We have shown the viability of our approach in GZSL and leave to future work the extension to other settings (e,g., few-shot or semi-supervised learning).

While our experimental results are encouraging, we argue that the strength of our formulation lies in its generality, and the full potential of PROTO-LTN is yet to be realized. Future work can aim at two complementary directions. First, alternative formulations of the \texttt{isOfClass} relationship could be explored, by changing the distance metric and/or the prototype encoding. Mapping class prototypes back to the input space, as done for instance in \cite{rudin2}, could improve explainability. 

Second, the knowledge $\mathcal{K}$ could be extended to leverage prior information, e.g., from external knowledge bases, to improve generalization to unseen classes. Experiments should include both inductive and transductive settings: the assumption that information about attributes and relationships of unseen classes is available at training or test time (e.g., from WordNet) is less restrictive than assuming that actual examples, albeit unlabelled, are available.

\bibliographystyle{IEEEtran}
\bibliography{main}

\onecolumn
\appendix[]

\subsection{Function grounding in PROTO-LTNs}
\label{sec:appendix-proto}

PROTO-LTNs are based on two functions ($\texttt{embeddingFunction}=f_\theta$ and $\texttt{getPrototype}$, respectively) and the $\texttt{isOfClass}$ predicate. 

The function  \texttt{getPrototypes}, with $\Din(\texttt{getPrototypes}) = \texttt{features} \times \texttt{labels}$ and \\ $\Dout(\texttt{getPrototypes}) = \texttt{embeddings} \times \texttt{labels}$,  returns labelled prototypes given a support set of labelled examples. In this way each prototype depends on the support set of the same class, as defined in Eq. \ref{batch_prototype}. As a consequence, we propose a novel definition for \textit{generalized} LTN functions.

To understand why a generalized function is needed, recall that LTN variables are grounded onto the set of their instantiations. Assume that $s$ is a variable associated to support points, or:
\begin{align*}
    \GG(s) = \braket{x^{\Tilde{S}}_1,...,x^{\Tilde{S}}_{N_S}}.
\end{align*}
If $h$ is a LTN function that is compatible with variable $s$, or $D_{\text{in}}(f)=D(s)=\reals^D$, the grounding for $h(s)$ is
\begin{align*}
    \GG(h(s)) = \braket{\GG(h)(x^{\Tilde{S}}_1),...,\GG(h)(x^{\Tilde{S}}_{N_S})}.
\end{align*}
This means that $\GG(h)$ only takes as input a single element of $\reals^D$. Unfortunately, a conventional LTN function such as $h$ cannot help us with prototypes, as their definition for a class $n \in \Tilde{C}$, given in Eq. \ref{batch_prototype}, is:
\begin{align*}
    p_n &= \frac{1}{K} \sum_{\substack{(x^{\Tilde{S}},y^{\Tilde{S}}) \in \Tilde{S} \\ \text{s.t. } y^{\Tilde{S}} = n}} f_\theta (x^{\Tilde{S}}) = p_n(x^{\Tilde{S}}_1,...,x^{\Tilde{S}}_{N_S}).
\end{align*}
Every prototype is in fact a function of all support points belonging to the same class. As a consequence, we propose a novel definition for \textit{generalized} LTN functions.

\begin{definition}
A generalized LTN function $F \in \FF^{\text{gen}}$ is a function that lets multiple instantiations of variables be fed at once to $\GG(F)$, and returns a variable.  The grounding for a generalized function $F \in \FF^{\text{gen}}$ is a function with flexible domain and range:
    \begin{align*}
        \GG(F): \bigcup_{l=1}^{\infty} \,
        \bigtimes_{m=1}^l \GG ( D_{\mathrm{in}}(F) ) \to \bigcup_{l=1}^{\infty} \, 
        \bigtimes_{m=1}^l \GG ( \Dout(F) ).
    \end{align*}
If a generalized function $F \in \FF^{\text{gen}}$ and a variable $x \in \XX$ have compatible domains, or $\Din(F) = D(x)$, the grounding for $F(x)$ is defined by
    \begin{align*}
        \GG(F(x)) = \GG(F)(\GG(x)).
    \end{align*}
\end{definition}

Grounding for both functions is defined as
\begin{align}
    \GG(\texttt{embeddingFunction}) &= f_\theta, \\
    \GG(\texttt{getPrototypes}) &= \Pi_\theta,
\end{align}
where $f_\theta: \reals^D \to \reals^M$  is the same embedding function as in the FSL setting, while
\begin{align*}
    \Pi_\theta: \bigcup_{l=1}^{\infty} \, \bigtimes_{m=1}^l \reals^D \times \naturals \to \bigcup_{l=1}^{\infty} \, \bigtimes_{m=1}^l \reals^M \times \naturals
\end{align*}

We structure $\Pi_\theta$ to be computationally easy to implement (e.g., in a computational graph), and to generalize to a setting in which $N_S$ and $\Tilde{N}$ are not fixed, or the $N$-way-$K$-shot scenario is not perfect. More specifically, in the following is how $\Pi_\theta$ works.
\begin{enumerate}
    \item Take as input:
    \begin{enumerate}
        \item a support set $\Tilde{S} = \{ (x^{\Tilde{S}}_1,y^{\Tilde{S}}_1), ..., (x^{\Tilde{S}}_{N_S},y^{\Tilde{S}}_{N_S}) \} \in (\reals^D \times \naturals)^{N_S}$ of labelled examples, with $x^{\Tilde{S}}_i \in \reals^D$ and $y^{\Tilde{S}}_i \in \naturals$;
        \item the parameter $\theta$ or, for the sake of clarity, the embedding function $f_\theta: \reals^D \to \reals^M$.
    \end{enumerate}
    \item Extract the classes contained in $\Tilde{S}$ by applying:
    \begin{align*}
        p^{(labels)} = \text{Unique}(y^{\Tilde{S}}),
    \end{align*}
    where the “Unique” function retrieves the unique elements of a vector. We call this variable $p^{(labels)}$ because it will be associated to prototype labels. Define $\Tilde{N}$ as the number of elements in $p^{(labels)}$.
    \item Define a sparse “labels” matrix $L \in \{0,1\}^{\Tilde{N} \times N_S}$ whose $i,j$-th entry is equal to 1 if support item $i$ is of class $p^{(labels)}_j$,  0 otherwise.
    \item Compute the prototypes tensor $p \in \reals^{\Tilde{N} \times M}$ as
    
    \begin{align*}
        \label{getprototypes}
        p = \text{Diag}(L \mathds{1}_{N_{S}})^{-1} \, L \, f_\theta(x^{\Tilde{S}})
    \end{align*}
    
    where
    \begin{align*}
        \mathds{1}_{N_S} = [1,1,...,1]^T \in \reals^{N_S}
    \end{align*}
    is a vector of $N_S$ ones, and
    \begin{align*}
        f_\theta(x^{\Tilde{S}}) = [f_\theta(x^{\Tilde{S}}_1), f_\theta(x^{\Tilde{S}}_2),...,f_\theta(x^{\Tilde{S}}_{N_S})]^T \in \reals^{N_S \times M}
    \end{align*}
    is the piece-wise application of $f_\theta$ to elements in $x^{\Tilde{S}}$, whereas “diag” computes the diagonal matrix associated to a vector. This expression does the same operation as Eq. \ref{batch_prototype}, but it is more general because it allows for unbalanced support sets. In the case of balanced support sets, which correspond to a perfect $N$-way-$K$-shot scenario, one simply has $\text{Diag}(L \mathds{1}_{N_S})^{-1} = \frac{1}{K} \, I$, where $I$ is the identity matrix.
    \item Return $p$ and $p^{(labels)}$.
\end{enumerate}

\subsection{Experimental settings details}
\label{sec:experiments-proto}
In Table \ref{tab:hyper} we report for each dataset the selected learning rate, $\alpha$ (distance parameter) and $\lambda$ (L2 regularization) which allowed us to obtain the best performance.

\setlength{\tabcolsep}{2pt} 
\renewcommand{\arraystretch}{-2} 
\begin{table*}[h!]
\centering
\caption{
Best hyperparameters used to train Proto-LTN on each benchmark. }
\label{tab:hyper}

  \begin{tabular}{l|SSS|}

    \toprule
    {Dataset} 
      & {Lr} & {$\alpha$} & {$\lambda$}    \\
      \midrule

     Awa2 &1e-04 & 1e-05 & 1e-03  \\

     CUB &1e-04 &1e-04 &1e-03 \\

     aPY&  1e-03&1e-05&1e-05    \\

     SUN & 1e-03&1e-05&1e-05  \\

    \bottomrule
  \end{tabular}
\end{table*}

\end{document}